\documentclass[letterpaper, 10 pt, conference]{ieeeconf}  

\IEEEoverridecommandlockouts                              

\usepackage{verbatimbox}
\usepackage{color}
\usepackage{footnote}
\usepackage{bbding}
\usepackage{algorithmicx,algorithm}
\usepackage{algpseudocode}
\usepackage{graphicx}
\usepackage{subfigure}
\usepackage{amsfonts}
\usepackage{booktabs}
\usepackage{threeparttable}
\usepackage{babel}
\usepackage{hyperref}
\hypersetup{
    colorlinks=true,
    linkcolor=blue,
    filecolor=blue,      
    urlcolor=blue,
    citecolor=cyan,
}

\usepackage{graphics} 
\usepackage{graphicx}
\usepackage{float}

\title{\LARGE \bf
Similarity-Aware Fusion Network for 3D Semantic Segmentation
}

\author{Linqing Zhao, Jiwen Lu and Jie Zhou
\thanks{Linqing Zhao is with the School of Mathematics, and the School of Electrical and Information Engineering, Tianjin University, Tianjin, 300072, China. Email: {\tt\small linqingzhao@tju.edu.cn}}%
\thanks{
Jiwen Lu is with the Department of Automation, Tsinghua University, Beijing, 100084, China. Email: {\tt\small lujiwen@tsinghua.edu.cn} (Jiwen Lu is the corresponding author of this paper.)}
\thanks{
Jie Zhou is with the Department of Automation, Tsinghua University, Beijing, 100084, China. Email: {\tt\small jzhou@tsinghua.edu.cn}}
\thanks{
Code is available at \url{https://github.com/lqzhao/SAFNet}}
}

\begin{document}

\maketitle
\thispagestyle{empty}
\pagestyle{empty}

\begin{abstract}

In this paper, we propose a similarity-aware fusion network (SAFNet) to adaptively fuse 2D images and 3D point clouds for 3D semantic segmentation.
Existing fusion-based methods achieve superior performances by integrating information from multiple modalities. However, they heavily rely on the projection-based correspondence between 2D pixels and 3D points and can only perform the information fusion in a fixed manner, so that their performances cannot be easily migrated to a more realistic scenario where the collected data often lack strict pair-wise features for prediction.
To address this, we employ a late fusion strategy where we first learn the geometric and contextual similarities between the input and back-projected (from 2D pixels) point clouds and utilize them to guide the fusion of two modalities to further exploit complementary information.
Specifically, we employ a geometric similarity module (GSM) to directly compare the spatial coordinate distributions of pair-wise 3D neighborhoods, and a contextual similarity module (CSM) to aggregate and compare spatial contextual information of corresponding central points.
The two proposed modules can effectively measure how much image features can help predictions, enabling the network to adaptively adjust the contributions of two modalities to the final prediction of each point. 
Experimental results on ScanNetV2 \cite{dai2017scannet} benchmark demonstrate that SAFNet outperforms existing state-of-the-art fusion-based approaches across various data integrity.

\end{abstract}

\section{INTRODUCTION}

3D semantic segmentation aims at predicting point-level annotations of different semantic categories for a given point cloud. It is a fundamental and crucial component of visual perception systems where autonomous robots work in a complex real-world environment composed of moving objects. It serves as the first procedure for a variety of downstream applications such as  autonomous mapping \cite{tateno20162,mahe2019real}, robot grasping \cite{satish2019policy,zhang2018road} and robot navigation \cite{rashed2019motion,teso2020semantic}. Since a single sensor cannot capture all necessary information, multiple types of sensors are often equipped on one robot in order to achieve more accurate segmentation results, where multi-modal fusion methods are needed to increase the capacity to compensate for disadvantages and rectify wrong predictions from each single sensor \cite{wu2019pointconv,liang2019multi,wang2019densefusion,Chen_2019_ICCV,qi2017frustum}.

\begin{figure}[t]
\includegraphics[scale=0.27]{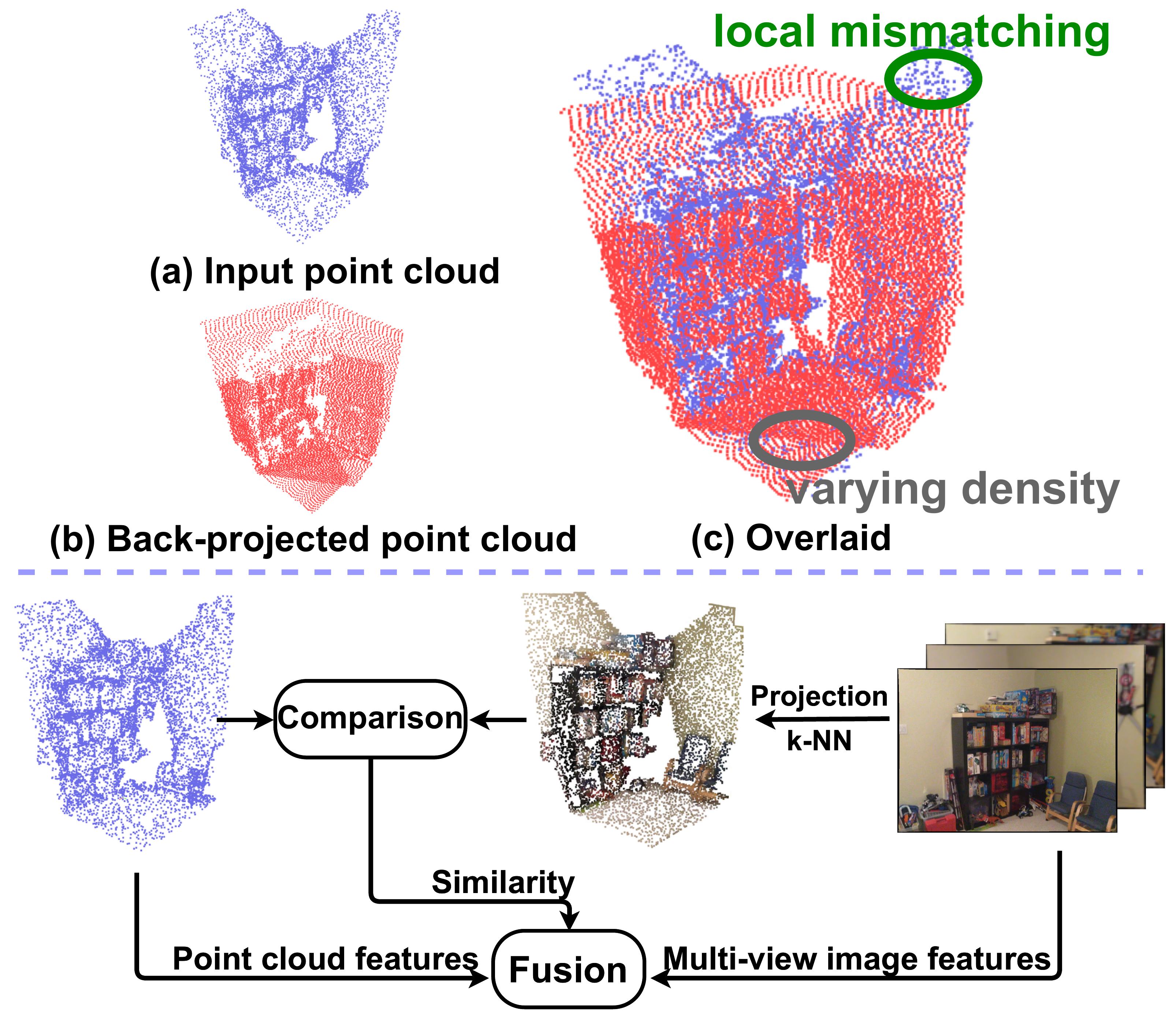}
\vspace{-7mm}
\caption{The upper part shows the comparison of \textbf{(a)} input point cloud, \textbf{(b)} back-projected point cloud and \textbf{(c)} an overlaid version in the same spatial range. We highlight two types of unsolved challenges within the green (\textbf{local mismatching}) and gray (\textbf{varying density}) ellipses. We downsample the point clouds uniformly at the same sample rate for a clear view.
The lower part is a high-level overview of our approach, where we use a comparison process to compute the similarity between the input and back-projected point clouds from different perspectives and further employ a fusion strategy to adaptively adjust the impact of the two modalities on classification results.} 
\vspace{-5mm}
\end{figure}

To learn a representation that takes the advantages of both modalities, 
existing works (e.g., 3DMV \cite{dai20183dmv}, UPB \cite{chiang2019unified}, MVPNet \cite{jaritz2019multi}, and FusionAwareConv \cite{zhang2020fusion}) have explored different ways to mine connections between point clouds and multi-view images, yet all of them extract the features of point clouds and multi-view images separately and fuse features in a fixed manner in a 3D space instead of the image plane. 
However, two unsolved issues (i.e., local mismatching \& varying density) limit the effectiveness of these methods in real scenes, as shown in the upper part of Fig. 1. 
Local mismatching is caused by the incoordination of multiple sensors and possible occlusions, making it hard for unmatched points to combine with suitable 2D appearance. 
Meanwhile, the overlap of multi-view images causes the density of back-projected point cloud to vary greatly in different spatial locations.

To address these, in this paper, we propose a similarity-aware late fusion framework to adaptively fuse image and point clouds, as shown in the lower part of Fig. 1. 
Firstly, we extract image features using a CNN for image segmentation and point features through a point-based network, which are high-dimensional informative representations of 2D appearance and 3D geometry, respectively.
Secondly, instead of directly concatenating features from two modalities, we propose to perform a comparison procedure between the input point cloud and back-projected point cloud (back-projected from multi-view images) to compute a similarity metric as the guidance of the following fusion process. 
We obtain this similarity metric by employing two modules to measure the per-point similarity of two point clouds from both the geometric and contextual perspective.
In addition, we employ a channel-wise attention layer for each modality to re-weight intra-modality channels prior to fusion to further boost performance.
At last, we adaptively combine the two types of features with the guidance of the aforementioned similarity metric.

The contributions of this paper can be summarized as:
\begin{itemize}

\item We propose a joint, end-to-end late fusion network aiming to infer 3D semantic segmentation from 3D point clouds and 2D images. To the best of our knowledge, this is the first 2D-3D fusion network able to handle various data integrity by tackling local mismatching and varying density.
\item We propose two efficient similarity modules to measure the geometric and contextual similarity of two point sets, which serve as a guidance to fuse image features and point cloud features. Note that this idea can be embedded in any 2D-3D fusion method.
\item  Our SAFNet outperforms previous published fusion-based methods by at least 1.3\% mIoU on ScanNetV2 benchmark across various data integrity, using the same backbone. Meanwhile, we provide an in-depth analysis of several ablation studies to demonstrate the effectiveness of our careful design.

\end{itemize}

\section{Related Work}

Currently, convolutional neural networks (CNN) based segmentation models have been the mainstream in 2D semantic segmentation. While in the case of 3D, methods can be categorized into point-based, multi-view and fusion-based methods. We give a brief review of these approaches.

\subsection{Point-based Methods}

 PointNet \cite{qi2017pointnet} is a notable landmark in the progress of point clouds processing \cite{guo2020deep} by deep learning approach, which leverages shared MLPs to learn per-point features and a max-pooling layer to obtain global features. However, PointNet fails to capture local structures, which makes it not suitable to handle large-scale point sets. To address this, PointNet++ \cite{qi2017pointnet++} presented a hierarchical architecture to learn local features with increasing contextual scales and improved the segmentation performance effectively. 
 KPConv \cite{thomas2019kpconv} defined a deformable convolution, making convolution kernel adaptable to local geometry and robust to varying densities. 
 Recently, several methods \cite{hu2019randla,yan2020pointasnl} have been proposed from the perspective of sampling points and aggregating information.
 However, existing point-based approaches only focus on the 3D geometry while ignoring the 2D clues that are complementary information to infer 3D labels.

\subsection{Multi-view Methods}

Multi-view CNNs are widely used in 3D reconstruction \cite{xie2019pix2vox}, \cite{yang2020robust},
3D shape retrieval \cite{su2015multi}, \cite{he2018triplet} which motivates several multi-view based works to deal with the 3D semantic segmentation problem. They usually predict labels in the 2D domain and then transfer them to 3D domain. For example, Hermans~\emph{et al.} \cite{hermans2014dense} classified RGB-D images using a Randomized Decision Forest and refined the results using a dense CRF before using a 2D-3D transfer approach to assign a class label to each 3D point.
SemanticFusion \cite{mccormac2017semanticfusion} used multi-view images to produce a SLAM map and a set of probability prediction maps, and then use a Bayesian update scheme to combine maps to obtain a final dense semantic map. Recently, TangentConv~\cite{Tatarchenko_2018_CVPR} built a U-type network using tangent convolutions as the main block for dense 3D segmentation.
Virtual-MVFusion \cite{kundu2020virtual} rendered synthetic images from meshes to train a 2D CNN and fused pixel predictions into 3D point predictions.
However, 2D information does not contain 3D geometry structure, which plays an important role in determining which label each point belongs to. Differently, our method leverages both 2D appearance and 3D geometry to jointly produce final predictions.

\begin{figure*}[t]
\centering
\includegraphics[scale=0.38]{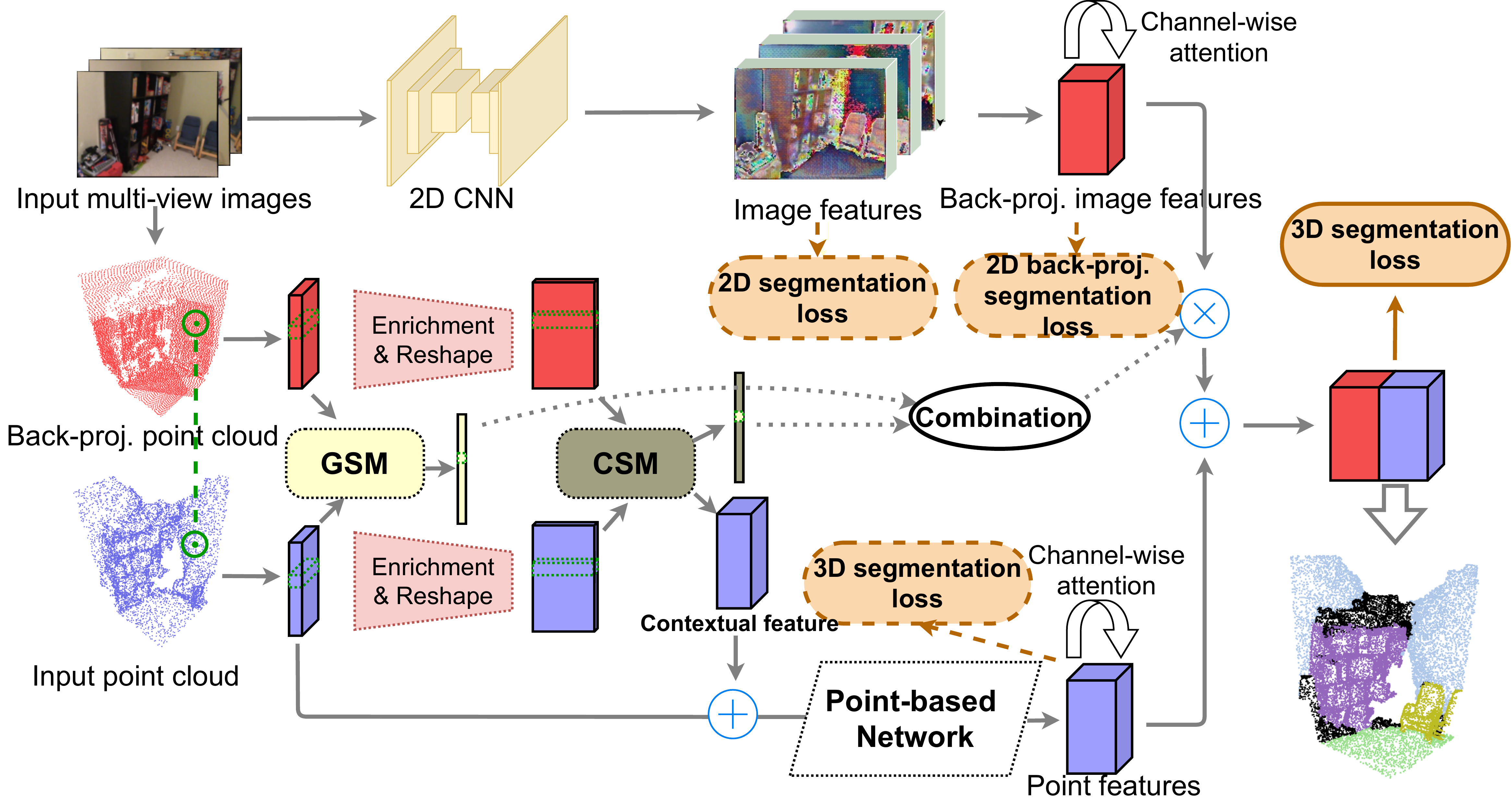}
\vspace{-8mm}
\caption{\textbf{The overall network architecture of the our proposed framework} can be divided into two branches (i.e., 3D branch and 2D branch) and a fusion header. For 3D branch, two neighborhoods centered on the same point pass through the geometric and contextual similarity module successively to obtain the enriched neighbor representation, as well as two similarities. Then, the mapped neighborhood representation by CSM is taken as attributes of points and fed into a point-based network to extract high-level point features.
For 2D branch, we abstract image  features via a fixed 2D encoder-decoder and back-project them into 3D canonical space. For fusion header, a high-efficiency strategy is proposed to combine features from two modalities under the guidance of geometric  and contextual similarity. Finally, fused features are fed into fully connected layers to predict the final semantic labels. } 
\vspace{-6mm}
\end{figure*}
\subsection{Fusion-based Methods}

Recently, 2D-3D fusion has been introduced in 3D segmentation task to enhance the robustness of systems and error handling. For example, 3DMV \cite{dai20183dmv} presented a joint late fusion architecture, where the features extracted from RGB images are back-projected into voxel volumes and combined with the features obtained from 3D branch to predict voxel labels. However, volumetric representation brings memory inefficiency and quantization error. UPB \cite{chiang2019unified} proposed a point-based early fusion architecture to learn 2D appearance, 3D structures and global context features from 3D meshes, which ensures all points to match with corresponding features from rendered images, avoiding the need for alignment. MVPNet \cite{jaritz2019multi} presented an early fusion framework that propagates image features to points by lifting dense pixel features to dense points features and then aggregating them into sparse points features. This strategy works effectively to fuse complementary information and achieves better results. However, it doesn't fully exploit the relationship between the input and the back-projected point cloud, which can serve as the mediator between 2D appearance and 3D geometry.

\section{Our Approach}

Compared with early fusion~\cite{zhang2020deep}, more fusion networks adopt late fusion structure in many tasks \cite{Liu_2013_CVPR,zhang2019late,8451915} due to the independence of feature extraction and convenience of interaction. However, the current best performing fusion method, MVPNet \cite{jaritz2019multi} is based on early fusion and we believe a late fusion structure with appropriate multi-modal interactions can achieve better performance and robustness.
Let $\mathcal{H}^{2D}(\theta^{2D})$, $\mathcal{H}^{3D}(\theta^{3D})$ be the 2D and 3D network with corresponding parameters $\theta^{2D}$ and $\theta^{3D}$. For input 2D image samples $\mathcal{X}^{2D}$ and 3D point samples $\mathcal{X}^{3D}$ with ground truth $\mathcal{Y}$, we denote the features extracted from two modalities as $\mathcal{F}^{2D} = \mathcal{H}^{2D}(\theta^{2D};\mathcal{X}^{2D})$ and $\mathcal{F}^{3D} = \mathcal{H}^{3D}(\theta^{3D};\mathcal{X}^{3D})$. For the traditional late fusion methods (e.g. 3DMV \cite{dai20183dmv}), the objective function can be written as:
\begin{equation}
\mathcal{L}_{fusion} = \mathcal{L}_{cls}(\mathcal{H}^{FC}([\mathcal{F}^{2D},\mathcal{F}^{3D}]);\mathcal{Y}),
\end{equation}
where $\mathcal{L}_{cls}$ is the widely used cross-entropy loss in segmentation tasks. $\mathcal{H}^{FC}$ represents a fully-connected network which maps the concatenation of $\mathcal{F}^{2D}$ and $\mathcal{F}^{3D}$ to the final prediction logits.

In this paper, we propose a similarity-aware fusion network and the architecture is shown in Fig. 2, where a per-point similarity is presented to realize multi-modal interaction by adjusting the contributions of two modalities to the final prediction. The proposed objective function can be written as:
\begin{equation}
\mathcal{L}_{fusion} = \mathcal{L}_{cls}(\mathcal{H}^{FC}([\Gamma(\mathcal{F}^{2D},\mathcal{S}^{2D-3D}),\mathcal{F}^{3D}]);\mathcal{Y}),
\end{equation}
where $\Gamma$ is a transformation operation to reweight 2D features using the similarity $\mathcal{S}^{2D-3D}$, which is learned from the geometric and contextual differences between input points and unprojected points (back-projected from image pixels). Note that we only reweight 2D features here for two reasons. On one hand, it is equivalent to adjusting the weights of both modalities. On the other hand, compared with 2D features, 3D features are more reliable and crucial for 3D segmentation so that 3D weights are fixed to accelerate convergence.
In the following sections, components of the network will be explained in detail.

\subsection{Image Feature Extraction and Back-projection}

In recent years, convolutional neural networks (CNN) have made great progress in understanding RGB images. To introduce informative 2D textures, a deep CNN is chosen as the 2D backbone. Since we cannot accurately process all points of the entire scene with a single forward, we follow the sliding window strategy used in PointNet++ \cite{qi2017pointnet++}. Hence, we need to select a number of RGB-D frames to cover as many points in each window as possible, where the depth channel is only used for back-projection. After selecting views, we feed the images into a pretrained 2D encoder-decoder network to obtain high-level feature maps $\mathcal{F}^{2D}$. Subsequently, pixels with features are back-projected into 3D space, utilizing depth values and camera matrixes, see Fig. 1. Accordingly, the back-projected points share the same coordinate system with the input points. It ensures that the pixel features of back-projected points can be gathered and combined with 3D geometric features from the corresponding input points in the future.

\subsection{Similarity Comparison Module}

In order to assign well suited pixel features to each point, it's necessary to distinguish some subsets of input points that need more or less appearance features. We argue that the point-wise similarity between input point set and back-projected point set can serve as the guidance information for reasonable fusion with image features. To this end, we present two effective similarity metrics to measure how similar two point sets are from the perspective of geometry and context, respectively. In this section, we elaborate on these two similarity module.

\begin{figure}[t]
\includegraphics[scale=0.455]{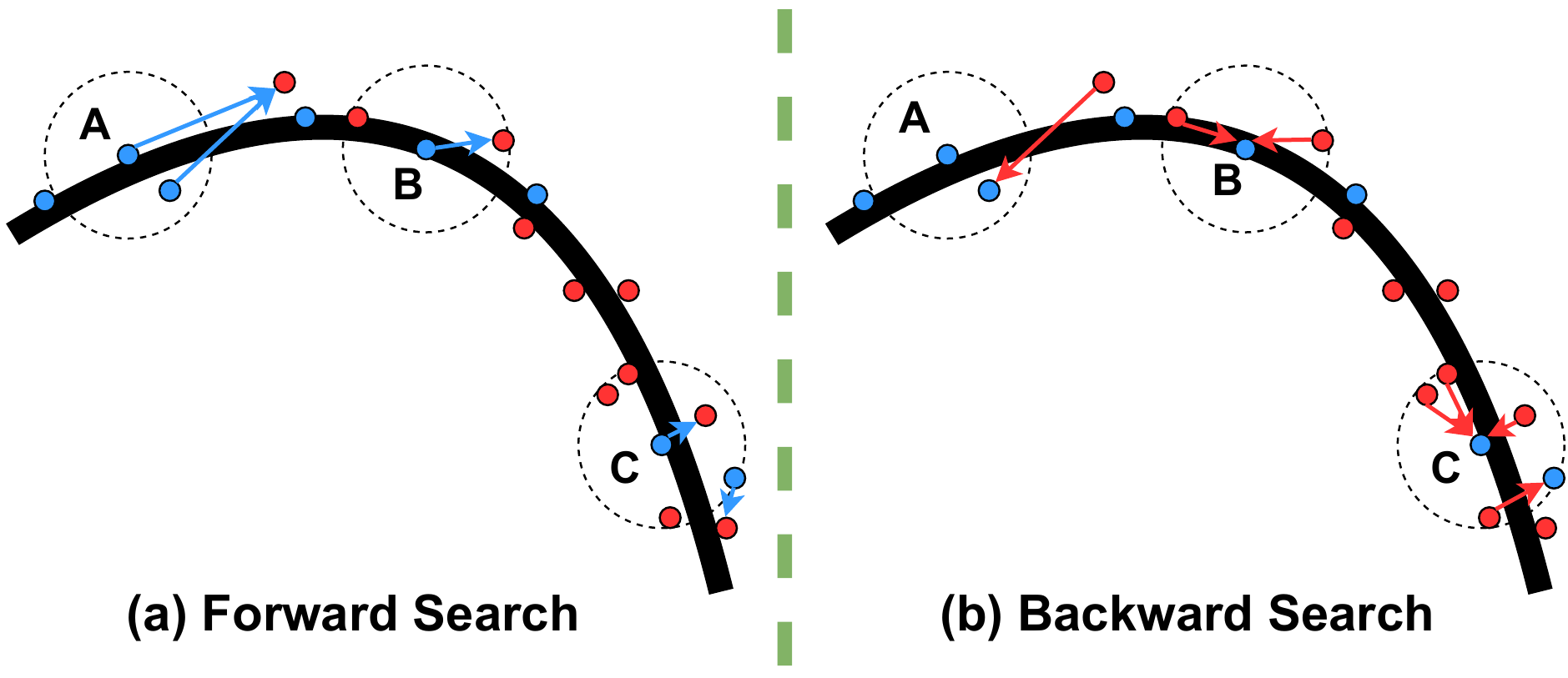}
\vspace{-7mm}
\caption{Illustration of our bidirectional nearest-neighbor distance search between input ($P$) and unprojected ($Q$) point clouds. 
The black curve represents a slice of a 3D surface on a 2D plane near which points from $P$ and $Q$ locate. Points A,B and C are from $P$, at the center of the corresponding neighborhood with different density. 
(a) \textbf{Forward search}: For each point in $N_{P,k}$, the search target is its nearest point in $Q$, which may not be in the neighborhood, like case A and C; (b) \textbf{Backward search}: For each point in $N_{Q,k}$, the search target is its nearest point in $N_{P,k}$. If $N_{Q,k}$ is empty like case A, the nearest point in $Q$ will be selected as the starting point.
Best viewed in color.} 
\vspace{-7mm}
\end{figure}

\subsubsection{\textbf{Geometric Similarity Module}}

For clarity and simplicity, we let $P = \left\{p_{1},p_{2},\ldots ,p_{n} \right\}$ denote original point clouds and $Q = \left\{q_{1},q_{2},\ldots ,q_{m} \right\}$ denote point clouds formed by back-projection, where $n$ and $m$ are the amount of points in two point sets, respectively. Note that $m$ is usually much greater than $n$, yet the vast majority of points in $Q$ actually are redundant due to their long distance to any point in $P$. So we only consider the neighborhood $N_{Q,k}(i)$, which consists of point $p_{i} \in P$ and its k-nearest neighbors within a radius $r$ in $Q$.

Generally, no matter which set points belong to, the points on the same spatial region should describe the same geometric shape. Hence, the low-level geometric representations of neighbor $N_{P,k}(i)$ and $N_{Q,k}(i)$, such as the distributions of 3D coordinates , should be highly similar, indicating a highly similar shape. However, the density, matching relation of points from two point sets can vary greatly across neighborhoods, which limits the effectiveness of shape comparison.

As shown in Fig. 3, the bidirectional nearest-neighbor distance mapping (BNNDM) between pair-wise neighbourhoods is proposed to directly measure the difference on geometry structures quantitatively. The concept of BNNDM consists of three parts: forward search, backward search and distance mapping. For forward search, we force each point in $N_{P,k}$ to find its nearest point in $Q$ and denote $d^{F}$  as the distance between them. Note that the found nearest point is not necessarily in the neighborhood in case of local mismatching. For backward search, each point in $N_{Q,k}$ is forced to find its nearest point in $N_{P,k}$ and denote $d^{B}$ as the distance. In the same way, we treat the case of mismatching specially by selecting the nearest point in $Q$ as the starting point. For distance mapping, we map $d^{F}$ and $d^{B}$ into a geometric similarity score $S_{G}$. Taking into account the negative correlation between distance and similarity, we consider to fit the relationship with negative exponential functions:
\begin{equation}
S^{P2Q}_i = exp(-\frac{1}{\alpha_1 n_i}\sum_{j=1}^{n_i}d^{F}_i)+\beta_1,                   p_j\in{\mathcal{N}}_{P,k}(i),
\end{equation}
\begin{equation}
S^{Q2P}_i =  exp(-\frac{1}{\alpha_2 m_i}\sum_{j=1}^{m_i}d^{B}_i)+\beta_2,                                     q_j\in{\mathcal{N}}_{Q,k}(i),
\end{equation}
\begin{equation}
S^{Geo}_i =  \alpha_3 S^{P2Q}_i + \alpha_4 S^{Q2P}_i + \beta_3,
\end{equation}
where $\alpha$ and $\beta$ are numerical variables, which can be adjusted to the optimal by network during training. Note that we also tried to use one-layer fully connected network with non-linear activation as an alternative mapping function but no improvement was brought on experimental results. 
The basic idea is twofold: on one hand, the average $d^{F}$ can represent distance of nearest point from Q, which is the key to determine whether there is a mismatch. On the other hand, the average $d^{B}$ is sensitive to the excess shape of Q, remaining unaffected by density changes of Q at the same time.
With such mapping function, geometric discrepancies, such as local mismatching and varying density, can be easily handled without redundancy learning parameters.

\subsubsection{\textbf{Contextual Similarity Module}}

\begin{figure}[t]

    \centering
    \includegraphics[scale=0.27]{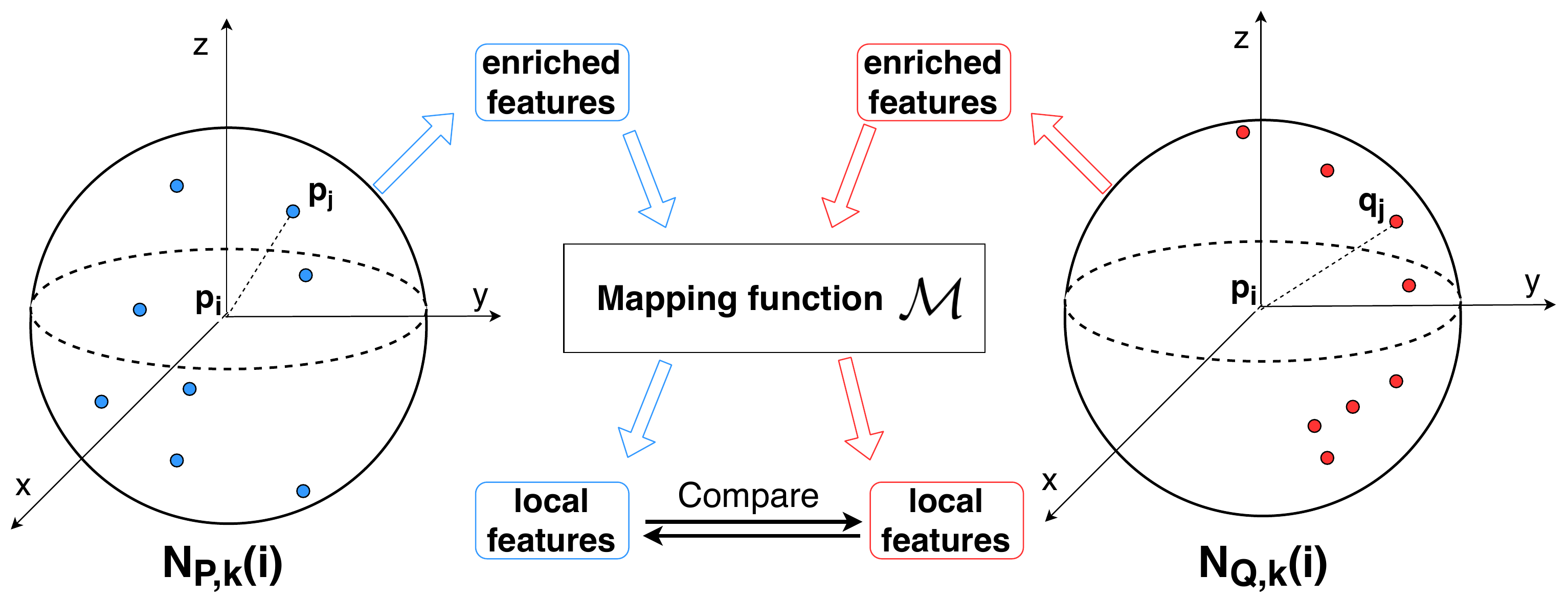}
    \vspace{-2mm}
    \caption{Illustration of point feature enrichment and contextual similarity comparison module. Neighborhood characters are be concatenated with the coordinates of central points to enrich point representations. Then, a following mapping operation convert enriched point representations to contextual features, from which the contextual similarity can be computed.} 
    \vspace{-5mm}
\end{figure}

Having obtained the geometric similarity of two point clouds, an intuitive idea is to regard it as the guidance of fusion. However, only low-level geometric similarity is insufficient to achieve the optimal fusion, since contextual clues are not fully utilized yet. 
To solve this, we present a contextual similarity module to compare high-level features of two point sets. Note that contextual features are no longer determined by only geometry, but also semantics.

Inspired by \cite{NIPS2019_8706}, we first enrich the point representation by augmenting additional neighborhood characters of each point.
Since encoding the distribution in eight directions \cite{jiang2018pointsift} helps to encode spatial information better, an orientation-encoding convolution layer followed by a shared multilayer perceptron (MLP) can act as the mapping function $\mathcal{M}$ to capture the local context. Therefore, for point $p_{i}$, the $C$-dimension neighborhood characters $f_{c}(x_{i},x_{j})$ and contextual feature $f_{i}$ can be presented as:

\begin{equation}
f_{c}(x_{i},x_{j}) = [x_{i}-x_{j},\Vert x_{i}-x_{j}\Vert^{2},...]
\end{equation}
\begin{equation}
f_{i} = \sum_{p_j\in{\mathcal{N}_{k}(i)}}\mathcal{\sigma}(w_{i,j}[x_{i},f_{c}(x_{i},x_{j})]+b_{i,j}),
\end{equation}
where $w_{i,j} \in \mathbb{R}^{C \times C}$ and $b_{i,j} \in \mathbb{R}^{C}$ are the trainable parameters in $\mathcal{M}$. $\sigma$ is the non-linear activation function. The illustration is shown in Fig. 4. Different from \cite{NIPS2019_8706}, our contextual information consists of pair-wise coordinate difference, Euclidean distance, etc. and will be fed into point-based network as affiliated attributes.
Given the contextual features $f^P_{i}$ and $f^Q_{i}$, we can easily conduct a cosine similarity function to measure how similar two neighborhoods are, which can be represented as:
\begin{equation}
S^{Con}_{i} = \frac{{(f^P_{i})^T  f^Q_{i}}}{\Vert f^P_{i}\Vert_{} \Vert f^Q_{i}\Vert_{}},
\end{equation}
where $T$ means vector transformation. Cosine similarity is extensively used in metric learning \cite{nguyen2010cosine,wojke2018deep} due to the special property that resulting similarity is always within [-1,1]. With contextual similarity,  a more in-depth comparison between two neighborhoods can be conducted through a lightweight design.

Finally, we feed all 3D coordinates $x \in P$ and their corresponding contextual feature $f$ into a point-based network to extract $\mathcal{F}^{3D}$, which can be represented as:
\begin{equation}
\{x,f\}\stackrel{P-Net}{\longrightarrow}{\mathcal{F}^{3D}}
\end{equation}

\begin{table*}[t]

\caption{Comparison with exising fusion-based methods on ScanNetV2 benchmark}
\vspace{1mm}
\begin{center}
\setlength{\tabcolsep}{0.9mm}{
\addvbuffer[-15pt -12pt]{
\begin{tabular}{cc|cccccccccccccccccccc}
\hline
Method & mIoU &bath&	bed&	bkshf&	cab&	chair	&cntr	&curt&	desk	&door	&floor	&other	&pic&	fridge&	shower&	sink&	sofa&	table&	toilet&	wall&	window\\
\hline
\hline
3DMV~\cite{dai20183dmv} & 48.4&48.4 &	53.8 &	64.3 &	42.4 &	60.6 &	31.0 &	57.4 &	43.3 &	37.8 &	79.6 &	30.1 &	21.4 &	53.7 &	20.8 &	47.2 &	50.7 &	41.3 &	69.3 &	60.2 &	53.9 \\
\hline
UPB~\cite{chiang2019unified} & 63.4& 61.4 &	\textbf{77.8}&	66.7 &	63.3&	\textbf{82.5}&	\textcolor{red}{\textbf{42.0}} &	\textbf{80.4}&	46.7 &	\textbf{56.1}&	\textbf{95.1}&	\textbf{49.4}&	29.1&	56.6 &	\textcolor{red}{\textbf{45.8}} &	57.9 &	\textbf{76.4}&	55.9 &	83.8 &	81.4&	59.8 \\
\hline
MVPNet~\cite{jaritz2019multi} & 64.1 & \textbf{83.1} &	71.5 &	\textcolor{red}{\textbf{67.1}} &	\textbf{59.0} &	78.1 &	39.4 &	67.9 &	\textbf{64.2 }&	\textcolor{red}{\textbf{55.3}} &	93.7 &	46.2 &	25.6 &	\textcolor{red}{\textbf{64.9}}&	40.6 &	62.6 &	69.1 &	\textbf{66.6} &	\textcolor{red}{\textbf{87.7}} &	79.2 &	60.8\\
\hline
FAConv~\cite{zhang2020fusion}&63.0&	60.4&	\textcolor{red}{\textbf{74.1}}&	\textbf{76.6}&	\textbf{59.0}	&74.7&	\textbf{50.1}&	73.4&	50.3&	52.7&	91.9&	45.4&	\textbf{32.3}&	55.0&	42.0&	\textbf{67.8}&	68.8&	54.4&	\textbf{89.6}&	\textbf{79.5}&	\textcolor{red}{\textbf{62.7}}\\
\hline
\hline
Ours & \textbf{65.4}	&\textcolor{red}{\textbf{75.2}}&	73.4&	66.4&	58.3&	\textcolor{red}{\textbf{81.5}}&	39.9&	\textcolor{red}{\textbf{75.4}}&	\textcolor{red}{\textbf{63.9}}&	53.5&	\textcolor{red}{\textbf{94.2}}&	\textcolor{red}{\textbf{47.0}}&	\textcolor{red}{\textbf{30.9}}&	\textbf{66.5}&	\textbf{53.9}&	\textcolor{red}{\textbf{65.0}}&	\textcolor{red}{\textbf{70.8}}&	\textcolor{red}{\textbf{63.5}}&	85.7&	\textcolor{red}{\textbf{79.3}}&	\textbf{64.2}\\
\hline
\end{tabular}
}}
\vspace{3mm}
\end{center}

\begin{tablenotes}
    \footnotesize
    \vspace{-1.5mm}
    \item[1] \rightline{* The \textbf{bold} and \textcolor{red}{\textbf{red}} values indicate the first-best and second-best results, respectively.}
    \vspace{-2.5mm}
    \item[2] \rightline{* All results are from published papers.}
    \vspace{-10mm}
\end{tablenotes}

\end{table*}


\begin{table}[t]
\caption{Comparison with multi-view and point-based methods on ScanNetV2 Benchmark}
\begin{center}
\setlength{\tabcolsep}{1mm}{
\begin{tabular}{cccc}
\hline
Method & mIoU &Input Modality&	Type\\
\hline
\hline

TangentConv~\cite{Tatarchenko_2018_CVPR}&40.9&point&multi-view\\
Virtual-MVFusion~\cite{kundu2020virtual}&74.6&mesh&multi-view\\
\hline
PN++~\cite{qi2017pointnet++}&33.9&point&point-based\\
PointSIFT~\cite{jiang2018pointsift}&41.5&point&point-based\\
PointCNN~\cite{li2018pointcnn}&45.8&point&point-based\\
DPC~\cite{9197503}&59.2&point&point-based\\
RandLA-Net~\cite{Hu_2020_CVPR} & 64.5&point&point-based\\
\hline
\hline
Ours (Res34\&PN++)&65.4&image+point&fusion\\
\hline
\end{tabular}
}
\vspace{-8mm}
\end{center}

\end{table}

\subsection{3D Fusion Network}

With the appearance and shape features (i.e. $\mathcal{F}^{2D}$ and $\mathcal{F}^{3D}$ ) of each point obtained from 2D and 3D branch, now we generate the fusion guidance through combining the geometric and contextual similarity calculated above (i.e. $S^{Geo}$ and $S^{Con}$).  Since labels should be predicted for input point cloud, we first attach dense image features to sparse points, by aggregating three nearest image  features , following the practice of \cite{jaritz2019multi}. 
Here we directly take the product of two similarities as the final weights, which can be represented by:
\begin{equation}
S^{2D-3D} = S^{Geo} \odot S^{Con},
\end{equation}
where $\odot$ represents the Hadamard product. 

To further improve the feature representation of specific semantics, we introduce the channel-wise attention \cite{fu2019dual} for both branches prior to fusion in order to model the interdependence between channels for each modality. 

Last but not the least, auxiliary supervision signals are indispensable for both modalities to accelerate convergence and make training process more stable. In addition to 2D and 3D direct supervision, an 2D back-projected semantic supervision is introduced to address 2D-3D inconsistent labels sometimes appear on the edges of objects. We achieve this by computing a cross entropy loss $\mathcal{L}_{2D_{unp.}}$ between 3D labels and nearest 2D back-projected features within a specific distance $r'$. Therefore, the total objective is composed of a fusion loss and three auxiliary loss:
\begin{equation}
\mathcal{L}_{toal} = \mathcal{L}_{fusion} + \lambda_1 \mathcal{L}_{2D} + \lambda_2 \mathcal{L}_{3D} + \lambda_3 \mathcal{L}_{2D_{unp.}},
\end{equation}
where $\lambda_1$, $\lambda_2$ and $\lambda_3$ are the coefficients of various loss components, and are set to 0.2, 0.8 and 0.8 in our implementation empirically. Here $\lambda_1$ is smaller because 2D CNN has been pretrained and only needs fine-tuning.

\section{Experiments}
To evaluate the effectiveness of the proposed approach, we conducted various experiments on the challenging ScanNetV2 dataset\cite{dai2017scannet}. We performed an ablation study to analyze the contribution of each similarity module, and the robustness to various data integrity. The evaluation metric is the mean intersection-over-union metric (mIoU). $IoU = \frac{TP}{TP+FP+FN}$, where $TP$, $FP$, and $FN$ are the numbers of true positive, false positive, and false negative points, respectively. Our experiments were conducted on the following specifications:
i7-4790K CPU, 32GB RAM, and NVIDIA GTX1080Ti GPU, using the PyTorch and Open3D \cite{Zhou2018} toolbox.

\subsection{ScanNetV2 Dataset}
ScanNetV2 is a large-scale richly annotated RGBD dataset of 1513 indoor scenes, which are provided with the RGB-D video sequences (depth-color aligned) and reconstructed meshes. There are 20 different scene types for the ScanNet dataset, such as apartments, bathrooms, conference rooms, including 40 types of indoor common objects. In our implementation, 1201 scenes were used for training and 312 scenes for testing. Since the ground truth of the test set is not available, our ablation study is based on the results of validation set.

\subsection{Implementation Details}

\subsubsection{\textbf{Pre-processing}}

To prevent spending much memory on image features, we first resized the resolution of RGBD images to 160x120. 
Since the network cannot handle too many points at once, we cut each scene into chunks of 1.5x1.5x3 $m^{3}$ to keep the average points in a chunk are about 25,000, following the practice of \cite{qi2017pointnet,qi2017pointnet++}.
Then, we back-projected image pixels utilizing camera intrinsic parameters, poses and depth maps to compute matching rates with all points of the corresponding scene. 
Finally, we selected a fixed number of frames for each chunk from consecutive video sequence to maximally cover points in the chunk with a greedy algorithm. The image that covers the most yet uncovered points is selected in every iteration.

\subsubsection{\textbf{Training 2D branch}}
We utilized the combination of ResNet34 \cite{he2016deep} and UNet \cite{ronneberger2015u} to extract 64-dimensional features as 2D appearance. We first initialized the CNN with weights pretrained on ImageNet and then we trained the CNN on ScanNetV2 2D Semantic dataset for 50000 iteration with the SGD optimizer from a learning rate of 0.005. We set the batch size to 84 and applied random cropping as well as horizontal random mirror for data augmentation. After training, we cut down the fully connected layers for classification and left the convolutional layers to provide image features in the overall network. Note that during point cloud network and fusion header training, parameters of CNN were fixed to speed up model convergence.

\subsubsection{\textbf{Training 3D branch}}
For both geometric and contextual similarity modules, we set the max number of points in a neighborhood, K, to 64. After passing through GSM, point representations were enriched and reshaped from (B,C,N,K) to (B,K(C+4),N) and fed into an orientation-encoding convolution and a three-layer-MLP of output dimensions (64,64,64), where there is a BN layer and ReLU after every MLP convolution layer. Another linear layer with 64 dimensional output are used to compute contextual features.
For point-based network, we used PointNet++ with single-scale grouping to get 128-dimensional 3D features. We set the input dimension of PointNet++ to 67 (3 for xyz and 64 for output of CSM).

\subsubsection{\textbf{Training fusion header}}
We set the output dimension of FC layers to (192,192,192,20), where there is a BN layer and ReLU after the first three layers and a dropout of 0.5 after the third layer to prevent overfitting.
For the whole 3D branch, we randomly sampled 8192 points in a chunk and trained the network on ScanNetV2 3D Semantic dataset using Adam optimizer with a learning rate of $10^{-3}$ (divided by 10 every 40 epochs) for 150 epochs.

\subsubsection{\textbf{Testing}}
For testing, we follow the practice of  \cite{qi2017pointnet++,jaritz2019multi}. The network predicts all the chunks with a stride of 0.5m in a sliding-window fashion. A majority vote is performed for the points located in multiple chunks.

\subsection{Results and Discussion}
\subsubsection{\textbf{Quantitative Results}}

We compared our method with recently proposed state-of-the-art methods on ScanNetV2 3D Semantic label benchmark. We employed the proposed SAFNet using 5 views with the chunk size of 1.5x1.5x3 $m^{3}$. For fair comparisons, all results listed here are provided by the published papers.

TABLE I shows the quantitative comparisons 
with exising fusion-based methods on the mean IoU and 20 different classes. Our proposed method achieves the best mIoU among state-of-the-art fusion-based approaches and top two performance of 70\% single class IoUs. In addition, our method achieves the smallest performance variance for 20 classes, which is an important basis for evaluating system stability. It’s worth noting that our 2D (i.e. ResNet34)and 3D  backbones (i.e. PointNet++) are the same as MVPNet's, but 
our mean IoU exceeds MVPNet by 1.3 \% without ensemble, which demonstrates the effectiveness of our similarity-aware fusion strategy. 
With the help of GSM and CSM, we learn a better fusion strategy to make use of the complementary information of two modailities.

We also conducted a comparison with state-of-the-art multi-view and point-based methods in TABLE III. As expected, our method outperforms most single-modality methods with a large margin. The insight is that fusion-based approaches can exploit both 2D appearance and 3D geometry to achieve joint prediction, compared with single-modality methods.
Although Virtual-MVFusion achieved superior results, it must take raw meshes as input in order to render much more images than ours from virtual perspectives, so that it can avoid the serious problems of mismatching and varying densities. In conclusion, the results well prove the effectiveness of our SAFNet in processing complete data.

\begin{figure}[t]
\centering
\subfigure{
	\includegraphics[width=1.55in]{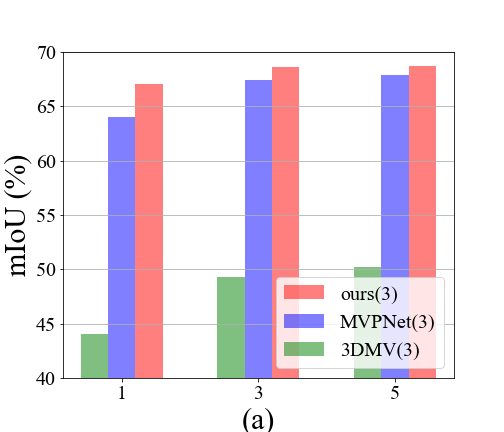}
}
\subfigure{
	\includegraphics[width=1.55in]{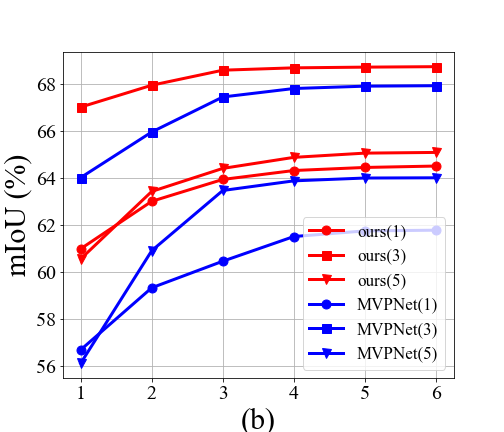}
}
\vspace{-3mm}
\caption{Comparison of robustness to the number of available  views for testing with the \textbf{(a)} same and \textbf{(b)} different number of views for training.
The format of the legends is \textit{method name(number of views for training)}.} 
\vspace{-6mm}
\end{figure}

Obtaining well-matched 2D images and 3D point clouds in practice is very expensive, so the robustness to various data integrity reflects the practicality of the system. 
Besides the complete data, SAFNet achieved impressive results on incomplete data, such as cases of fewer images. 
We compared with fusion-based 3DMV\cite{dai20183dmv} and MVPNet \cite{jaritz2019multi} with the same number of views for training in Fig. 5 (a). We also conducted a detailed comparison with MVPNet with various numbers of views for training in Fig. 5 (b).
For a fair comparison, we used the same  training settings for all models and the same backbones as MVPNet. Note that UPB \cite{chiang2019unified} uses images rendered from meshes to keep every point matching with image pixels so that we don't compare with UPB here. Clearly, our models with different training settings consistently outperform their counterparts. As decreasing available views for testing, the gap between our method and the others grows wider. In addition, our performance curves in Fig. 5 (b) are much smoother, which demonstrates that our method is more robust to data incompleteness. Interestingly, the training strategy with 3 views available achieves the best performance across different testing cases, compared with the other two training strategies. We conjecture that 3 views can provide enough well-matched pixel-voxel pairs for training 2D and 3D deep networks as well as some unmatched cases for training GSM and CSM. On the contrary, too few well-matched pairs (1 view) would result in under-fitting while too few unmatched cases (5 views) would limit the generalization ability.

\begin{figure}[t]
\centering
\includegraphics[scale=0.18]{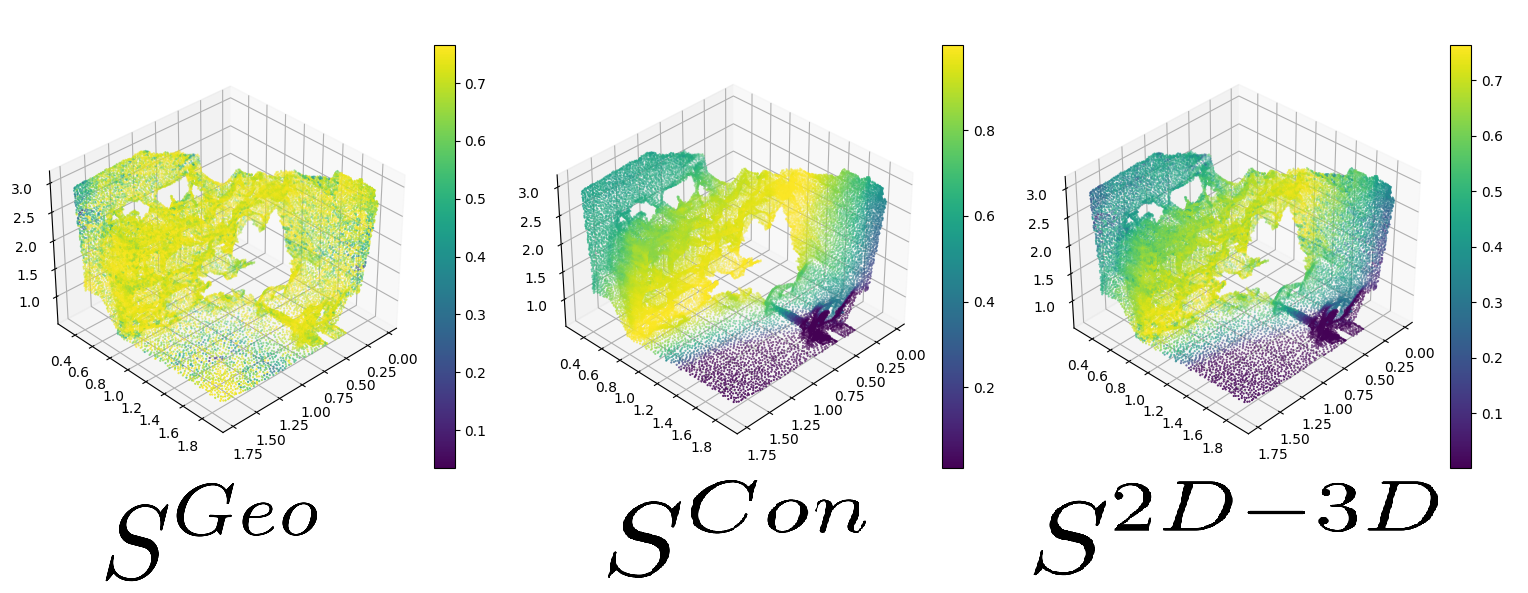}
\vspace{-3mm}
\caption{Example of learned geometric similarity ($S^{Geo}$), contextual similarity ($S^{Con}$) and the combination ($S^{2D-3D}$).}
\vspace{-3mm}
\end{figure}

\begin{figure}[t]
\centering
\includegraphics[scale=0.15]{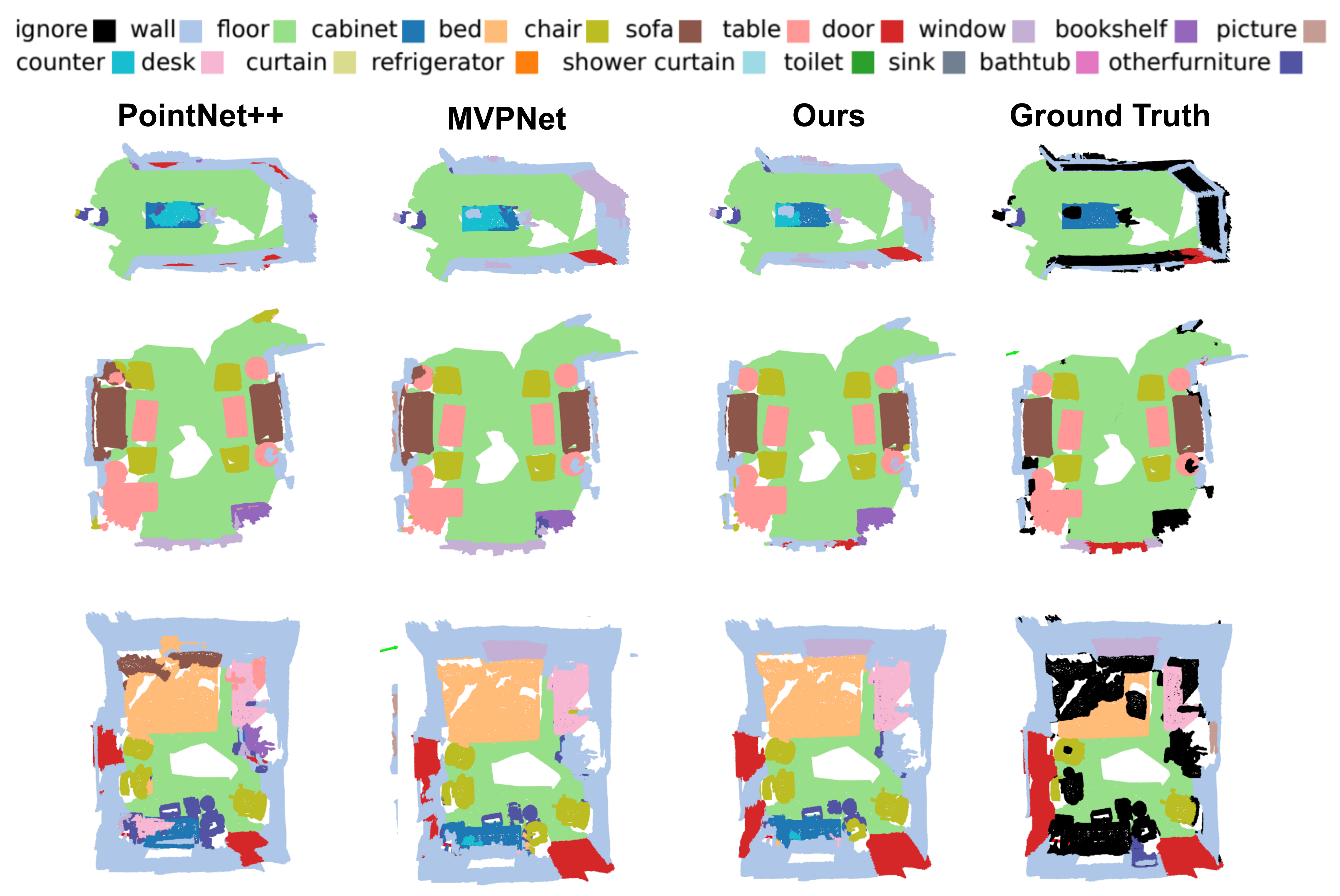}
\vspace{-4mm}
\caption{Quantitive results of 3D segmentation on the validation set of ScanNetV2.}
\vspace{-7mm}
\end{figure}

\subsubsection{\textbf{Qualitative Results}}
Fig. 6 shows an example of learned similarities, which can effectively guide the fusion of two modalities.
Fig. 7 shows the segmentation results of our method and other two state-of-the-art methods on the validation set of ScanNetV2.

\subsubsection{\textbf{Discussion}}
Experimental results show that our method achieves very competitive results on both complete and incomplete data. We conclude the insights as follows:
\begin{itemize}

\item Our GSM and CSM can efficiently capture geometric and contextual information, which indicates matching condition, density changing, situation of occlusion and other properties while previous methods didn't take them into consideration or only consider the distance from a point to its k-nearest neighbors in point set $Q$, $N_{Q,k}$. Unfortunately, for the case of mismatches, the points in $N_{Q,k}(i)$ are usually too far away from the center point $p_i$, so that the image features they provide are mostly wrong. 
\item We can adaptively adjust the impact ratio of features from images and point cloud in different areas while previous works simply combine them in a fixed manner, which results in suboptimal predictions even in well-matched areas.

\end{itemize}
\subsection{Ablation Studies}
In this section, we conduct ablation studies on the validation set of ScanNetV2  to investigate the contribution of individual components in the SAFNet. In order to control variables,  the number of input views was fixed to 3 for both training and testing. For clarity, we consider the baseline model: ResNet34-UNet (2D) + PointNet++ (3D) + FC (header) and we separately analyzed each component as follows:

\begin{table}[tb]
    \centering
    \caption{Ablation studies on the validation set of ScanNetV2 
    }
    \renewcommand\tabcolsep{4.5pt}

    \begin{tabular}{c|c c c c c c c c c}
      \hline  
      \textbf{Component} & \multicolumn{8}{|c}{\textbf{Design Choice}} \\
      \hline
            Baseline  &  \Checkmark  & \Checkmark &\Checkmark &\Checkmark&\Checkmark&\Checkmark&\Checkmark  &   \Checkmark\\
      GSM-FS  &   & \Checkmark &&\Checkmark&\Checkmark&\Checkmark&\Checkmark  &   \Checkmark\\
      GSM-BS&  & &\Checkmark&\Checkmark&\Checkmark&\Checkmark&\Checkmark&   \Checkmark \\
      CSM&  & &&&\Checkmark&\Checkmark&\Checkmark &   \Checkmark\\
      Auxiliary Sup   &   &            &          &&&\Checkmark&\Checkmark &   \Checkmark \\
      2D Unp. loss    &   &            &          &          &&       &   \Checkmark  &   \Checkmark           \\
      Channel Att    &   &            &          &          &          &&    &   \Checkmark    \\
 
      \hline
  
      mIoU(\%)           & 61.5  & 63.5      &62.8 & 64.2     & 66.5     & 67.9     & 68.3     & 68.5    \\ 
      \hline
     \end{tabular}
     \vspace{-7mm}
\end{table}

\subsubsection{\textbf{Effect of Geometric similarity module}}

We split GSM into Forward Search (GSM-FS) and Backward Search (GSM-BS) to compare with baseline, as shown in TABLE II. We can summarize three points from the comparison. In the first place, the model with GSM-FS achieves 2\% performance improvement while GSM-BS can boost 1.3\% over the baseline, which demonstrates both parts of GSM can effectively capture 3D geometry and map it to instructive fusion weights. Second, Forward Search can provide more direct guidance to fusion since it's more sensitive to local mismatch. At last, the combination of two parts can further boost performance, which means both parts extracted some unique information.

\subsubsection{\textbf{Effect of Contextual similarity module}}

In the experiments, we compared the models with or without CSM. For the model without CSM, we directly feed the 3D coordinates into PointNet++ to extract high-dimension features. The significant performance gap between two models proves the effectiveness of CSM. We argue that extracting compact neighborhood representation before feeding into deep networks can be helpful in many tasks, especially segmentation and detection.

\subsubsection{\textbf{Loss functions}}

We conducted two experiments to investigate the effectiveness of auxiliary supervision and 2D unprojected loss. As shown in TABLE II, the model with 2D and 3D auxiliary supervision achieves 1.4\% gain, compared with only 3D supervision at the end of network and this strategy can provide constraints for intermediate features, making network training more stable. In addition, the small gain obtained from 2D unprojected loss demonstrates that the proposed unprojected loss can address 2D-3D inconsistent labels. It can prevent the network from getting confused by inconsistent labels.

\subsubsection{\textbf{Channel-wise attention}}

Applying channel-wise attention to the two branches at the same time can indirectly realize the interaction of the two modalities. Experimental results show that channel-wise attention provides 0.2\% gain.

\section{CONCLUSIONS}

In this paper, we have proposed a new framework to fuse 2D images and 3D point clouds by computing image features and point features individually and then fusing them with the help of two learned similarities from the perspectives of geometry and context. Comprehensive experiments have been conducted on ScanNetV2 Semantice segmentation benchmark, and the experimental results demonstrate our superior performance as well as the robustness to various data integrity. This method can be used in autonomous mapping, robot grasping, and robot navigation. 

\section{Acknowledgement}

We thank Prof. Zhanjie Song for supporting me in life and learning. We also thank Liangliang Ren and Wenzhao Zheng for their valuable advice.
This work was supported in part by the National Key Research and Development Program of China under Grant 2017YFA0700802, National Natural Science Foundation of China under Grant 61822603, Grant U1813218, Grant U1713214.






\bibliographystyle{ieeetr} 
\bibliography{ref} 
\addtolength{\textheight}{-12cm}   
\end{document}